# On Nonspecific Evidence

Johan Schubert
*Division of Applied Mathematics and Scientific Data Processing,*
*Department of Weapon Systems, Effects and Protection, National*
*Defence Research Establishment, S-172 90  Sundbyberg, Sweden*

When simultaneously reasoning with evidences about several different events it is necessary to separate the evidence according to event. These events should then be handled independently. However, when propositions of evidences are weakly specified in the sense that it may not be certain to which event they are referring, this may not be directly possible. In this paper a criterion for partitioning evidences into subsets representing events is established. This criterion, derived from the conflict within each subset, involves minimising a criterion function for the overall conflict of the partition. An algorithm based on characteristics of the criterion function and an iterative optimisation among partitionings of evidences is proposed.   © 1993 John Wiley & Sons, Inc.

## I.   INTRODUCTION

A problem of major importance when reasoning with uncertainty is that in many situations evidences will not only be uncertain but their propositions may also be weakly specified in the sense that it may not be certain to which event a proposition is referring. In some cases the propositions may not carry any such information, making it impossible to differentiate between different events. Furthermore, the domain knowledge regarding events may be uncertain. For instance, our knowledge of the current number of events may only be probabilistic.

When reasoning about some proposition it is crucial not to combine evidences about different events in the mistaken belief that they are referring to the same event. For this reason every proposition's action part must be supplemented with an event part describing to which event the proposition is referring. The event part may be more or less weakly specified dependent on the evidence. If the evidences could be clustered into subsets representing events that should be handled separately from the others the situation would become manageable.

A simple example will illustrate the terminology. Let us consider the burglaries of two bakers' shops at One and Two Baker Street, event 1 ($E_1$) and event 2 ($E_2$), i.e., the number of events is known to be two. One witness hands over an evidence, specific with respect to event, with the proposition: "The burglar at One Baker Street," event part: $E_1$, "was probably brown haired ($B$)," action part: $B$. A second anonymous witness hands over a nonspecific evidence with the proposition: "The burglar at Baker Street," event part: $E_1$, $E_2$, "might have been red haired ($R$)," action part: $R$. That is, for example:

| evidence 1: | evidence 2: |
|---|---|
| proposition: | proposition: |
| action part: $B$ | action part: $R$ |
| event part: $E_1$ | event part: $E_1, E_2$ |
| $m(B) = 0.8$ | $m(R) = 0.4$ |
| $m(\Theta) = 0.2$ | $m(\Theta) = 0.6$ |

The aim of this paper is to establish, within the framework of Dempster–Shafer theory,[1–3] a criterion function[4] of overall conflict when reasoning with multiple events. With this criterion we may handle evidences whose proposition is weakly specified in its event part. We will use the minimizing of overall conflict as the method of partitioning the set of evidences into subsets representing the events. This method will also handle the situation when the number of events are uncertain.

An algorithm for minimizing the overall conflict will be proposed. The proposed algorithm is based on the one hand on characteristics of the criterion function for varying number of subsets and on the other hand on an iterative optimization among partitionings of evidence for a fixed number of subsets.

This algorithm was developed as a part of a multiple-target tracking algorithm for an antisubmarine intelligence analysis system.[5,6] In this application a sparse flow of intelligence reports arrives at the analysis system. These reports carry a proposition about the occurrence of a submarine at a specified time and place, a probability of the truthfulness of the report and may contain additional information such as velocity, direction and type of submarine.

The intelligence reports are never labeled as to which submarine they are referring to but it is of course possible to differentiate between two different submarines of two intelligence reports if the reports are known to be referring to different types of submarines. Moreover, times and positions of two different reports may be such that it is impossible to travel between the two positions at their respective times and therefore possible to differentiate between the two submarines. However, when this is not the case differentiation will not be possible. Instead we will use the conflict between the two intelligence reports as a probability that the two reports are referring to different submarines.

Before analysing the possible tracks for an unknown number of submarines we want to separate the intelligence reports into subsets according to which submarine they are referring to and then analyse the possible tracks for each submarine separately. The most probable partition of reports into subsets is done by minimizing the criterion function of overall conflict.

In this application the action part of the intelligence report proposition states that a submarine was at the indicated time and position while the event part of a report is informal, often weakly specified and contained in the information that to some degree separates reports, such as time and position of the other reports, nonfiring sensors placed between reports, etc.

In Sec. II we will discuss how to use the overall conflict for separating nonspecific evidences. Following this, the criterion function for overall conflict of multiple events will be investigated (Sec. III). We then discuss the behaviour of the criterion function in iterative optimization (Sec. IV). Finally, we propose an algorithm for partitioning evidences into subsets (Sec. V), based on the criterion function and a hill-climbing−like iterative optimization. We conclude with a detailed example (Sec. VI).

## II.   SEPARATING NONSPECIFIC EVIDENCE

When we have evidences with conflicting event parts we would like to separate them into disjoint subsets. After this, the reasoning should take place with the evidences in each subset treated separately. However, when the event part of a proposition is weakly specified with respect to which of many different events it is referring, it may be difficult if not impossible to directly judge whether or not it and a second proposition are referring to the same event. If, for instance, the first proposition is referring to events one or two and the second proposition is referring to events two or three it is uncertain whether or not they are referring to the same event. Thus, it will not be possible to separate evidences based only on their proposition's event parts.

Instead we will separate evidences by their conflict. This is an obvious choice since the conflict measures the lack of compatibility among evidences and the action parts of propositions are more likely compatible when they are referring to the same event as compared to the situation when they are referring to different events where the actions are also most likely different. Evidences are considered conflicting when they have empty intersections between representations of the proposition action parts with identical specific proposition event parts, i.e. propositions certainly referring to one and the same event. However, since all calculations take place within subsets where the evidences are presumed to be referring to the same event, we will have a conflict in two different situations. Firstly, we have a conflict if the proposition action parts are conflicting regardless of the proposition event parts since they are presumed to be referring to the same event. Secondly, if the proposition event parts are conflicting then, regardless of the proposition action parts, we have a conflict with the presumption that they are referring to the same event. In order to avoid that evidences with specific and identical event parts end up in different subsets we may precombine these evidences and henceforth handle them as one evidence. The idea of using the conflict as distance measure between bodies of evidence has been suggested earlier by Lowrance and Garvey[7] and by Lesh.[8]

The conflict within each subset will not only be seen as a measure of the lack of compatibility among evidences within the subset but also as an evidence against the current partitioning of the set of evidences, $\chi$, into the subsets, $\chi_i$. This is an intuitively correct definition since a critique against a part of the partitioning, the lack of compatibility among evidences, is a critique against the entire partitioning, i.e. an evidence against the partitioning. A zero conflict is no evidence against the partitioning and a conflict of one is an evidence that the partitioning is impossible. The frame of discernment is here $\Theta = \{$ Partition, $\neg$Partition$\}$. That is, the basic probability assignment against the partitioning from subset $\chi_i$ is

$$m_{\chi_i}(\neg\text{Partition}) \overset{\Delta}{=} \text{Conf}(\{e_j | e_j \in \chi_i\}),$$

$$m_{\chi_i}(\Theta) \overset{\Delta}{=} 1 - \text{Conf}(\{e_j | e_j \in \chi_i\})$$

where $\{e_j | e_j \in \chi_i\}$ is the set of evidences belonging to subset $\chi_i$ and $\text{Conf}(\cdot)$ is the conflict, $k$, in Dempster's rule.

When the evidences are not simple support functions the conflict measure might, at first glance, seem odd as a distance measure between bodies of evidence, since two nonsimple support functions with identical sets of focal elements may have a nonzero conflict. However, this need not be nonintuitive, as shown by the case of four simple support functions, the two first identical and in conflict with the two identical remaining simple support functions. If the two first and the two last functions are combined then a conflict measure with the intuitive properties of a distance measure is obtained, when the two resulting simple support functions are combined. If, on the other hand, we combine the first with the third and the second with the fourth, we receive two identical nonsimple support functions whose combination will result in a nonzero conflict. Clearly, if the conflict measure was intuitive as distance measure in the first combination order then it is also intuitive in the second. Then, at least for support functions that are derivable from simple support functions, it is not nonintuitive to have a nonzero distance measure for support functions with identical sets of focal elements.

In addition there will also be a domain dependent conflict from a probability distribution about the number of subsets, $E$, conflicting with the actual current number of subsets, $\#\chi_i$. This conflict will also be seen as an evidence against the current partitioning of the set of evidences into the subsets,

$$m_D(\neg\text{Partition}) \overset{\Delta}{=} \text{Conf}(\{E_i\}, \#\chi_i),$$

$$m_D(\Theta) \overset{\Delta}{=} 1 - \text{Conf}(\{E_i\}, \#\chi_i).$$

Fusing these evidences with Dempster's rule yields

$$m(\neg\text{Partition}) = 1 - [1 - m_D(\neg\text{Partition})] \cdot \prod_{i=1}^{r} \left[1 - m_{\chi_i}(\neg\text{Partition})\right].$$

$$m(\Theta) = [1 - m_D(\neg\text{Partition})] \cdot \prod_{i=1}^{r} \left[1 - m_{\chi_i}(\neg\text{Partition})\right]$$

with belief and plausibility of the partitioning being

$$\text{Bel(Partition)} = 0,$$

$$\text{Pls(Partition)} = [1 - m_D(\neg\text{Partition})] \cdot \prod_{i=1}^{r} \left[1 - m_{\chi_i}(\neg\text{Partition})\right].$$

Finding the most probable partitioning of evidences into disjoint subsets representing different events will then be the problem of maximizing the plausibility of possible partitionings, or the dual problem of minimizing one minus the plausibility. The difference, one minus the plausibility of a partitioning will be called the metaconflict of the partitioning.

## III.  METACONFLICT AS A CRITERION FUNCTION

Let us define the metaconflict function, derived in the previous section, whose minimization per definition leads to the optimal partioning of evidences into disjoint subsets.

DEFINITION. *Let the* metaconflict function,

$$Mcf(r, e_1, e_2, ..., e_n) \stackrel{\Delta}{=} 1 - (1 - c_0) \cdot \prod_{i=1}^{r} (1 - c_i), \tag{1}$$

*be the conflict against a partitioning of n evidences of the set $\chi$ into r disjoint subsets $\chi_i$ where*

$$c_0 = \sum_{i \neq r} m(E_i) \tag{2}$$

*is the conflict between r subsets and propositions about possible different number of subsets and*

$$c_i = \sum_{I| \cap I = \varnothing} \prod_{e_j^k \in I} m(e_j^k)$$

*is the conflict in subset i, where $I = \{e_j^k | e_j \in \chi_i\}$  is a set of one focal element from the support function of each evidence in $\chi_i$.*

Some characteristics of the metaconflict function will be useful when choosing the number of subsets of $\chi$ for which we must find an optimal partitioning of evidences.

The first theorem below states that if we have an optimal partitioning for *r*

subsets, then we need never consider any solutions with fewer than $r$ subsets when the basic probability number for $r$ subsets is greater than the basic probability number for fewer subsets. These solutions need never be considered because the nondomain part of the metaconflict function always increases with fewer subsets and when the basic probability number for fewer subsets is smaller than the basic probability number for $r$ subsets, then the domain part of the metaconflict function for fewer subsets has also increased, yielding an overall increase in the metaconflict. The significance of this theorem is that it can be applied iteratively. If we first find the optimal partitioning for the number of subsets where $m(E_r)$ is greatest, we need never consider any solutions with fewer subsets than $r$, and if we then find the optimal partitioning for the greatest $m(E_j)$ where $j > r$, then we need never consider any further solutions where the number of subsets are fewer than $j$, etc.

The second theorem states that if we have an optimal partitioning for some number of subsets we need never consider any solutions for some other number of subsets where the domain part of the metaconflict function is greater than the metaconflict of our present partitioning. This theorem will also be used iteratively as we gradually find better optimizations, step by step eliminating some of the possible solutions where the number of subsets is greater than with our present partitioning.

Together, these two theorems will significantly reduce the number of iterative optimizations we must carry through for different numbers of subsets.

THEOREM 1. *For all $j$ with $j < r$, if $m(E_j) < m(E_r)$ then min Mcf($r$, $e_1$, $e_2$, ..., $e_n$) < min Mcf($j$, $e_1$, $e_2$, ..., $e_n$).*

*Proof.* From the fact $m(E_j) < m(E_r)$ and (2) it follows

$$c_0 = \sum_{i \neq r} m(E_i) = \sum_{i \neq j} m(E_i) + m(E_j) - m(E_r) < \sum_{i \neq j} m(E_i) = c'_0 . \qquad (3)$$

From (3) and by the definition of metaconflict (1) it is sufficient that

$$\forall j . \max \prod_{i=1}^{j} (1 - c'_i) \leq \max \prod_{i=1}^{r} (1 - c_i)$$

for min Mcf($r$, $e_1$, $e_2$, ..., $e_n$) to be less than min Mcf($j$, $e_1$, $e_2$, ..., $e_n$). This is equivalent with

$$\forall j . \max \prod_{i=1}^{j} (1 - c'_i) \leq \max \prod_{i=1}^{j+1} (1 - c_i) .$$

It is sufficient to show that the partition into $j$ subsets that yields the maximum is less than or equal to any partition into $j + 1$ subsets.

Let the partition into the $j$ first of the $j + 1$ disjoint subsets be unchanged from the optimal partition into $j$ subsets with the exception that one evidence is moved from one of the subsets with more than one evidence, say $\chi_k$, to subset $\chi_{j+1}$. There is, with only one evidence, no conflict in $\chi_{j+1}$, $c_{j+1} = 0$.

Then

$$\prod_{i=1}^{j+1} (1 - c_i) = \prod_{i=1}^{j} (1 - c_i) = (1 - c_k) \cdot \prod_{\substack{i=1 \\ \neq k}}^{j} (1 - c_i)$$

$$= (1 - c_k) \cdot \prod_{\substack{i=1 \\ \neq k}}^{j} (1 - c_i') = \frac{1 - c_k}{1 - c_k'} \cdot \prod_{i=1}^{j} (1 - c_i') \geq \prod_{i=1}^{j} (1 - c_i')$$

since the conflict in $\chi_k$ after moving out an evidence , $c_k$, is always less than or equal to the conflict before moving, $c_k'$ .

THEOREM 2. *For all j, if min Mcf(r, $e_1$, $e_2$, ..., $e_n$) < $\sum_{i \neq j} m(E_i)$ then min Mcf(r, $e_1$, $e_2$, ..., $e_n$) < min Mcf(j, $e_1$, $e_2$, ..., $e_n$).*

*Proof.* From the condition of the theorem and by (1) we have

$$\min Mcf(r, e_1, e_2, ..., e_n) < \sum_{i \neq j} m(E_i)$$

$$= \frac{\min Mcf(j, e_1, e_2, ..., e_n) + \prod_{i \neq j} (1 - c_i) - 1}{\prod_{i \neq j} (1 - c_i)}$$

$$= 1 - \frac{1 - \min Mcf(j, e_1, e_2, ..., e_n)}{\prod_{i \neq j} (1 - c_i)}$$

$$< \min Mcf(j, e_1, e_2, ..., e_n).$$

There are also two theorems regarding the stability of an optimal solution, i.e. that the partion of the optimal solution can not self-splinter into new subsets, Theorem 3, and that the partition is invariant with respect to evidence incompatible with the partition, Theorem 4.

Since the nondomain part of the metaconflict function decreases with the number of subsets it is only the domain conflict part of the metaconflict that prevents the number of subsets to be equal to the number of evidences. Thus, whether or not a partitioning of evidences is stable depends on the relation between these conflicts.

THEOREM 3. *A partitioning is stable, i.e. the metaconflict increases if any evidence is removed from its subset to form a new subset, if the relative change in domain conflict is higher than all relative conflict changes of the subsets.*

*Proof.* If an evidence $e_q$ is removed from $c_i$ and included into a new subset $\chi_{r+1}$ the metaconflict would change to

$$Mcf^* = 1 - (1 - c_0^*) \cdot (1 - c_i^*) \cdot \prod_{k \neq i} (1 - c_k)$$

$$= 1 - (1 - c_0) \cdot (1 - c_i') \cdot \prod_{k \neq i} (1 - c_k)$$

$$-[\,(1-c_0^*)\cdot(1-c_i^*)-(1-c_0)\cdot(1-c_i)\,]\cdot\prod_{k\neq i}(1-c_k)$$

$$=Mcf+[\,(c_0^*-c_0)+(c_i^*-c_i)-(c_0^*\cdot c_i^*-c_0\cdot c_i)\,]\cdot\prod_{k\neq i}(1-c_k)$$

The partition is stable if $\forall i$ . $Mcf^* > Mcf$. That is, if

$$\forall i.\frac{(c_0^*-c_0)}{1-c_0}>\frac{(c_0\cdot c_i-c_0^*\cdot c_i^*)-(c_i^*-c_i)}{1-c_0}$$

$$>c_i^*-c_i>\frac{(c_i^*-c_i)}{1-c_i}.$$

Finally, a new evidence which is uncompatible with all subsets and introduced into its own subset will not change the partition.

THEOREM 4. *If $P$ is a unique optimal partition of the set of evidences, $\chi$, and $e_{n+1}$ a new evidence which is highly conflicting with each subset $\chi_i$ is introduced into its own subset $\chi_{r+1}$. Then the optimal partition of $\chi \cup \{e_{n+1}\}$ is $P / \{ e_{n+1} \}$.*

*Proof.* The optimal partition of the new set of evidences is found by minimizing $Mcf^*(r + 1, e_1, e_2, \ldots, e_{n+1})$. However, since $e_{n+1}$ is introduced into its own subset $\chi_{r+1}$, a subset without conflict, this can be rewritten as a function of the minimization of the old metaconflict, $Mcf(r, e_1, e_2, ..., e_n)$;

$$\min Mcf^*(r + 1, e_1, e_2, ..., e_{n+1})$$

$$=\min 1-(1-c_0^*)\cdot\prod_{i=1}^{r+1}(1-c_i)$$

$$=\min 1-(1-c_0^*)\cdot\prod_{i=1}^{r}(1-c_i)$$

$$=\min 1-\frac{1-c_0^*}{1-c_0}\cdot(1-c_0)\cdot\prod_{i=1}^{r}(1-c_i)$$

$$=\min 1-\frac{1-c_0^*}{1-c_0}\cdot[1-Mcf(r,e_1,e_2,...,e_n)]$$

$$=1-\frac{1-c_0^*}{1-c_0}\cdot[1-\min Mcf(r,e_1,e_2,...,e_n)]\,.$$

That is, finding the optimal partition of $\chi \cup \{e_{n+1}\}$ when $e_{n+1}$ is introduced into its own subset is done by finding an optimal partition of $\chi$. Since there is only one such partition, $P$, the optimal partition of $\chi \cup \{e_{n+1}\}$ is $P / \{e_{n+1}\}$.

## IV. CONDITION FOR ITERATIVE OPTIMIZATION

For a fixed number of subsets a minimum of the metaconflict function can be found by an iterative optimization among partitionings of evidences into different subsets. This approach is proposed in order to avoid the combinatorial problem in minimizing the metaconflict function. In each step of the optimization the consequence of transferring an evidence from one subset to another is investigated. If an evidence $e_q$ is transferred from $\chi_i$ to $\chi_j$ then the conflict in $\chi_j$, $c_j$, increases to

$$c_j^* = c_j + \sum_{\substack{A_k \in \chi_j | A_k \neq \varnothing \\ e_q^p \in e_q | A_k \cap e_q^p = \varnothing}} m(e_q^p) \cdot m(A_k).$$

with new focal elements and basic probability assignments

$$m^*(A_k) = m(A_k) \cdot \sum_{e_q^p \in e_q | A_k = A_k \cap e_q^p} m(e_q^p)$$

and

$$m^*(A_k \cap e_q^p) = m(A_k) \cdot \sum_{e_q^p \in e_q | A_k \neq A_k \cap e_q^p} m(e_q^p)$$

where $\{A_k\}$ are the focal elements before the transfer of $e_q$ and $\{A_k, e_q^p \cap A_k\}$ are the focal elements after the transfer. The conflict in $\chi_i$, $c_i$ decreases to

$$c_i^* = c_i - \sum_{\substack{A_k \in \chi_i | A_k \neq \varnothing \\ e_q^p \in e_q | A_k \cap e_q^p = \varnothing}} m(e_q^p) \cdot m^*(A_k)$$

where

$$m^*(A_k) = m(A_k) \Big/ \sum_{e_q^p \in e_q | A_k = A_k \cap e_q^p} m(e_q^p)$$

Here, $\{A_k\}$ are the focal elements after the transfer of $e_q$ and $\{A_k, e_q^p \cap A_k\}$ are the focal elements before the transfer. That is, we find the basic probability assignment of the focal elements as if the evidence was not included in $\chi_i$ and calculate the additional conflict created by transferring the evidence to $\chi_i$. This additional conflict is then deducted from the conflict, $c_i$, to calculate the conflict after having transferred the evidence from $\chi_i$, $c_i^*$. Given this, the metaconflict is changed to

$$Mcf^* = 1 - (1 - c_0) \cdot (1 - c_i^*) \cdot (1 - c_j^*) \cdot \prod_{k \neq i, j} (1 - c_k)$$

$$= 1 - (1 - c_0) \cdot \prod_k (1 - c_k)$$

$$+ (1-c_0) \cdot \left( \prod_k (1-c_k) - (1-c_i^*) \cdot (1-c_j^*) \cdot \prod_{k \neq i,j} (1-c_k) \right)$$

$$= Mcf + (1-c_0) \cdot [ \, (1-c_i) \cdot (1-c_j)$$

$$- (1-c_i^*) \cdot (1-c_j^*) \, ] \cdot \prod_{k \neq i,j} (1-c_k).$$

The transfer of $e_q$ from $\chi_i$ to $\chi_j$ is favourable if Mcf* < Mcf. From the last expression, this is the case if

$$(1-c_i) \cdot (1-c_j) < (1-c_i^*) \cdot (1-c_j^*).$$

Rewriting this as

$$\frac{1-c_j^*}{1-c_j} > \frac{1-c_i}{1-c_i^*}$$

we substitute $c_i^*$ and $c_j^*$ with their expressions

$$\frac{1-c_j - \displaystyle\sum_{\substack{A_k \in \chi_j | A_k \neq \varnothing \\ e_q^p \in e_q | A_k \cap e_q^p = \varnothing}} m(e_q^p) \cdot m(A_k)}{1-c_j} > \frac{1-c_i}{1-c_i + \displaystyle\sum_{\substack{A_k \in \chi_i | A_k \neq \varnothing \\ e_q^p \in e_q | A_k \cap e_q^p = \varnothing}} m(e_q^p) \cdot m^*(A_k)}$$

which yields

$$1 - \frac{\displaystyle\sum_{\substack{A_k \in \chi_j | A_k \neq \varnothing \\ e_q^p \in e_q | A_k \cap e_q^p = \varnothing}} m(e_q^p) \cdot m(A_k)}{1-c_j} > \frac{1}{1 + \dfrac{\displaystyle\sum_{\substack{A_k \in \chi_i | A_k \neq \varnothing \\ e_q^p \in e_q | A_k \cap e_q^p = \varnothing}} m(e_q^p) \cdot m^*(A_k)}{1-c_i}}.$$

Finally, we conclude that the transfer of $e_q$ from $\chi_i$ to $\chi_j$ is favourable if

$$\frac{\displaystyle\sum_{\substack{A_k \in \chi_j | A_k \neq \varnothing \\ e_q^p \in e_q | A_k \cap e_q^p = \varnothing}} m(e_q^p) \cdot m(A_k)}{1-c_j} < \frac{\dfrac{\displaystyle\sum_{\substack{A_k \in \chi_i | A_k \neq \varnothing \\ e_q^p \in e_q | A_k \cap e_q^p = \varnothing}} m(e_q^p) \cdot m^*(A_k)}{1-c_i}}{1 + \dfrac{\displaystyle\sum_{\substack{A_k \in \chi_i | A_k \neq \varnothing \\ e_q^p \in e_q | A_k \cap e_q^p = \varnothing}} m(e_q^p) \cdot m^*(A_k)}{1-c_i}}.$$

Let us call these quotients $\rho_j^q$ and $\rho_i^q$ respectively, i.e., it is favourable to transfer $e_q$ from $\chi_i$ to $\chi_j$ if $\rho_j^q < \rho_i^q$. It is, of course, most favourable to transfer $e_q$ to $\chi_k$, $k \neq i$, if $\forall j . \rho_k^q \leq \rho_j^q$. It should be remembered that this analysis concerns the situation where only one evidence is transferred from one subset to another. It may not be favourable at all to simultaneously transfer two or more evidences which are deemed favourable for individual transfer. It can easily be shown that when several different evidences are favourable to transfer it will be most favourable to transfer the evidence $e_q$ that maximizes $(1 - \rho_k^q)/(1 - \rho_i^q)$.

## V.   AN ALGORITHM FOR MINIMIZING METACONFLICT

The algorithm for finding the partitioning of evidences among subsets that minimizes the metaconflict is based on Theorems 1 and 2 of the metaconflict function for finding the optimal number of subsets and an iterative optimization among partitionings of evidences for a fixed number of subsets. The iterative part of the algorithm, step 4 in the algorithm below, guarantees, like all hill-climbing algorithms, local but not global optimum.

**Algorithm.** Let $S$ be the set of natural numbers less or equal to the number of evidences and $T$ the empty set.

1. *Calculate* $\forall r. \sum_{i \neq r} m(E_i)$, *the conflict against a partitioning of the evidences into $r$ subsets.*
2. *Let* $r = j | \min_{j \in S} \sum_{i \neq j} m(E_i)$
3. $T = T + \{r\}, S = S - \{r, j | j < r, j \in S\}$
4. *Calculate min Mcf($r$, $e_1$, $e_2$, ..., $e_n$).*
   4.1. *Make an initial partition equal to the final partition of the last calculation of Mcf into the first $t$ subsets with the exception of moving the $r - t$ most highly conflicting evidences from these subsets, updating the conflicts after each movement, one into each of the new $r - t$ subsets. If it is the first calculation make any partition with at least one evidence in each subset. Calculate Mcf of the current partition.*
   4.2. *Let* $t = r$. *If* $r = 1$ *go to 4.5.*
   4.3. *For* $q = 1$ *to* $n$. *Suppose that* $e_q$ *is currently in* $\chi_i$.
      4.3.1. *IF* $|\chi_i| = 1$ *go to 4.3 else calculate for* $1 \leq j \leq r$,

$$\rho_j^q = \begin{cases} \dfrac{\displaystyle\sum_{\substack{A_k \in \chi_j | A_k \neq \varnothing \\ e_q^p \in e_q | A_k \cap e_q^p = \varnothing}} m(e_q^p) \cdot m(A_k)}{1 - c_j}, & j \neq i \\[4em] \dfrac{\dfrac{\displaystyle\sum_{\substack{A_k \in \chi_i | A_k \neq \varnothing \\ e_q^p \in e_q | A_k \cap e_q^p = \varnothing}} m(e_q^p) \cdot m^*(A_k)}{1 - c_i}}{1 + \dfrac{\displaystyle\sum_{\substack{A_k \in \chi_i | A_k \neq \varnothing \\ e_q^p \in e_q | A_k \cap e_q^p = \varnothing}} m(e_q^p) \cdot m^*(A_k)}{1 - c_i}}, & j = i \end{cases}$$

4.4. *Transfer $e_q$ from $\chi_i$ to $\chi_k$, $k \neq i$, if*

$$\forall w. \frac{1 - \rho_k^q}{1 - \rho_i^q} \geq \frac{1 - \rho_k^w}{1 - \rho_i^w}$$

*where* $\forall j. \rho_k^w \leq \rho_j^w$.

4.5. *Update Mcf, $c_i$ and $c_k$.*

4.6. *If Mcf is unchanged in step 4.5 then go to 5 else go to 4.3.*

5. $S = S - \left\{ \begin{array}{l} \forall j \mid j > r \\ \left| \sum_{i \neq j} m(e_i^\chi) > min \quad Mcf(r, e_1, e_2, ..., e_n) \right. \end{array} \right\}$.

6. *If $S \neq \varnothing$ then go to 2 else answer $min_{t \in T} Mcf(t, e_1, e_2, ..., e_n)$.*

## VI. AN EXAMPLE

Let us, as an illustration of the problem solved in this paper, consider a simple example of two possible burglaries, with a couple of evidences with simple support functions and some of the evidence weakly specified in the sense that it is uncertain to which possible burglary their propositions are referring. Assume that a baker´s shop at One Baker Street has been burglarized, event 1. Let there also be some indication that a baker´s shop across the street, at Two Baker Street, might have been burglarized, although no burglary has been reported, event 2. An experienced investigator estimates that a burglary has taken place at Two Baker Street with a probability of 0.4. We have received the following evidences. A credible witness reports that "a brown-haired man who is not an employee at the baker´s shop committed the burglary at One Baker Street," evidence 1. An anonymous witness, not being aware that there might be two burglaries, has reported "a brown-haired man who works at the baker´s shop committed the burglary at Baker Street," evidence 2. Thirdly, a witness reports having seen "a suspicious-looking red-haired man in the baker´s shop at Two Baker Street," evidence 3. Finally, we have a fourth witness, this witness, also anonymous and not being aware of the possibility of two burglaries, reporting that the burglar at the Baker Street baker´s shop was a brown-haired man. That is, for example: elds:

*evidence 1:*
  *proposition:*
    *action part: BO*
    *event part:* $E_1$
$m(BO) = 0.8$
$m(\Theta) = 0.2$

*evidence 2:*
  *proposition:*
    *action part: BI*
    *event part:* $E_1$, $E_2$
$m(BI) = 0.7$
$m(\Theta) = 0.3$

*evidence 3:*
  *proposition:*
    *action part: R*
    *event part:* $E_2$
$m(R) = 0.6$
$m(\Theta) = 0.4$

*evidence 4:*
  *proposition:*
    *action part: B*
    *event part:* $E_1$, $E_2$
$m(B) = 0.5$
$m(\Theta) = 0.5$

*domain probability distribution:*

$$m(E_i) = \begin{cases} 0.6, i = 1 \\ 0.4, i = 2 \\ 0, i \neq 1, 2 \end{cases}.$$

Let us use the algorithm in Sec. V to investigate whether we have a one or two event problem and possibly separate the set of evidences into two disjoint subsets.

**Algorithm:** Let $S = \{1, 2, 3, 4\}$, $T = \varnothing$, where $S$ are possible numbers of subsets and $T$ different numbers of subsets for which we have minimized Mcf.

Step 1: We calculate the domain conflict from the probability distribution,

$$c_0 = \begin{cases} 0.4, r = 1 \\ 0.6, r = 2 \\ 1, r \neq 1, 2 \end{cases}.$$

Step 2: The domain conflict is minimal for one subset, $r = 1$.

Step 3: Update $T := T + \{1\} = \{1\}$, $S := S - \{1\} = \{2, 3, 4\}$.

Step 4:

    Step 4.1: In the initial partition all evidences are brought into one subset $\chi_1 = \{e_1, e_2, e_3, e_4\}$. Mcf = 0.884.

    Step 4.2: $t = 1$. Since we only have one subset, $r = 1$, no transfers are possible, we go to 4.5.

    Step 4.5: No evidences has been transferred, Mcf and $c_1$ is unchanged.

    Step 4.6: Since Mcf was unchanged in step 4.5 we go to 5.

Step 5: We update, $S := S - \{3, 4\} = \{2\}$.

Step 6: Since $S \neq \varnothing$ there might exist better solutions, we go to 2.

Step 2: We minimize Mcf for two subsets, $r = 2$.

Step 3: Update, $T := T + \{2\} = \{1, 2\}$, $S := S - \{2\} = \varnothing$.

Step 4:

    Step 4.1: As the initial partitioning move the most highly conflicting evidence from $\chi_1$ to $\chi_2$. We have $e_1$: $\rho_1^1 = 0.604$, $e_2$: $\rho_1^2 = 0.578$, $e_3$: $\rho_1^3 = 0.559$, $e_4$: $\rho_1^4 = 0.085$, i.e. we move $e_1$ from subset $\chi_1$ to subset $\chi_2$. Mcf = 0.804.

    Step 4.2: $t = 2$.

    Step 4.3: Since $e_1$ is in $\chi_2$ and $|\chi_2| = 1$, $e_1$ can not be moved out of $\chi_2$ and no $\rho_j^i$'s are calculated. For $q = \{2, 3, 4\}$ we get for, $e_2$: $\rho_1^2 = 0.3$, $\rho_2^2 = 0.56$, $e_3$: $\rho_1^3 = 0.51$, $\rho_2^3 = 0.48$, $e_4$: $\rho_1^4 = 0.155$, $\rho_2^4 = 0$.

    Step 4.4: We get for, $e_2$: $(1 - \rho_2^2)/(1 - \rho_1^2) = 1$, $e_3$: $(1 - \rho_2^3)/(1 - \rho_1^3) = 1.061$, $e_4$: $(1 - \rho_2^4)/(1 - \rho_1^4) = 1.184$, i.e. we move $e_4$ from $\chi_1$ to $\chi_2$. We get $\chi_1 = \{e_2, e_3\}$ and $\chi_2 = \{e_1, e_4\}$.

    Step 4.5: We update conflicts, Mcf = 0.768, $c_1 = 0.42$, $c_2 = 0$.

    Step 4.6: Mcf has changed and we continue at 4.3.

    Step 4.3: Since $|\chi_1| = |\chi_2| = 2$ all evidences can be moved. For $q = \{1, 2, 3, 4\}$ we get for, $e_1$: $\rho_1^1 = 0.634$, $\rho_2^1 = 0$, $e_2$: $\rho_1^2 = 0.42$, $\rho_2^2 = 0.48$, $e_3$: $\rho_1^3 = 0.42$, $\rho_2^3 = 0.54$, $e_4$: $\rho_1^4 = 0.155$, $\rho_2^4 = 0$.

Step 4.4: For all $q$ we have $\rho_k^q = \rho_i^q$, i.e. no evidences are transferred.

Step 4.5: Mcf, $c_1$ and $c_2$ are unchanged.

Step 4.6: Since Mcf is unchanged, go to 5.

Step 5: S := Ø, there are no further possible solutions.

Step 6: Since S = Ø we answer $\{\chi_1, \chi_2, ..., \chi_t\}$ where $t \in T = \{1, 2\}$ minimizes the metaconflict function,

$$\min_{t \in T} Mcf(t, e_1, e_2, ..., e_n),$$

i.e. we answer $\{\chi_1, \chi_2\}$ where $\chi_1 = \{e_2, e_3\}$, $\chi_2 = \{e_1, e_4\}$ for $t = 2$. Concluding, we see from the event parts of the evidences in each subset that $\chi_1$ corresponds to event 2 and $\chi_2$ corresponds to event 1.

## VII.    CONCLUSIONS

A criterion function of overall conflict has been established within the framework of Dempster–Shafer theory. An algorithm has been proposed for partitioning nonspecific evidence into subsets, each subset representing a separate event. The algorithm has a theoretical foundation in the minimizing of overall conflict of the partition when viewing the conflict within each subset as an evidence against the partition. The algorithm will not only be able to reason about the optimal partition of nonspecific evidence for a fixed number of events, it will also be able to reason simultaneously about the optimal number of events, which may be uncertain. An obvious drawback is the algorithm's inability to guarantee global optimality

I would like to thank Stefan Arnborg, Ulla Bergsten and Per Svensson for their helpful comments regarding this article